\documentclass[10pt,twocolumn,letterpaper]{article}

\usepackage{cvpr}              
\usepackage{adjustbox}
\usepackage{multirow}
\usepackage{colortbl}   
\usepackage{cuted}
\definecolor{cvprblue}{rgb}{0.21,0.49,0.74}
\usepackage[pagebackref,breaklinks,colorlinks,allcolors=cvprblue]{hyperref}

\def\paperID{xxx} 
\def\confName{CVPR}
\def\confYear{2026}

\title{VVitCutLER: Towards Unsupervised Object Detection and Segmentation in Videos}

\author{Zhijing Lu$^{1,2}$ \quad
Khurram Azeem Hashmi$^{2,1}$ \quad
Didier Stricker$^{2,1}$ \quad
Muhammad Zeshan Afzal$^{2,1}$ \\
$^{1}$RPTU University of Kaiserslautern-Landau, Germany\\
$^{2}$German Research Center for Artificial Intelligence, Germany\\
{\tt\small zhijinglu011@gmail.com}
}

\begin{document}

\maketitle

\begin{strip}
\includegraphics[width=0.95\textwidth]{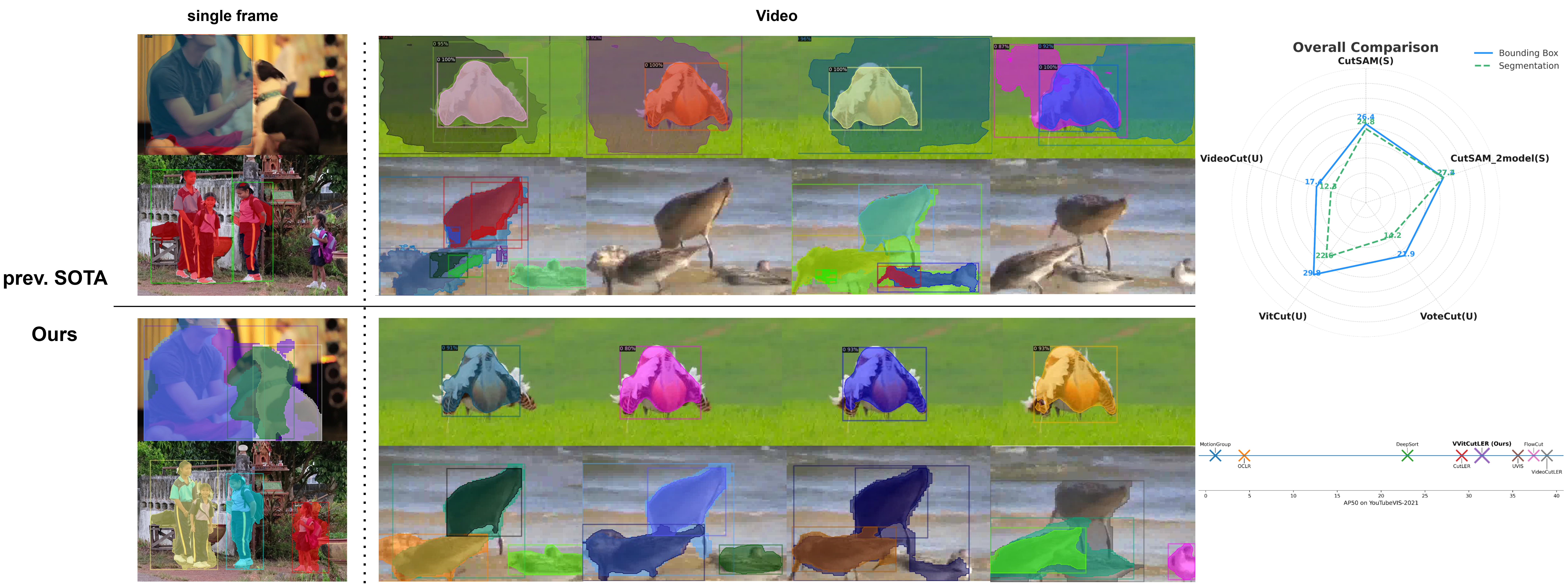}
    \vspace{-5pt}
    \captionof{figure}{
        \textbf{Unsupervised object detection and instance segmentation.}
        The single-frame comparison (left) shows a comparison between our annotation module VitCut (used in the preprocessing stage of VVitCutLER) and the reference method VoteCut. VitCut generates significantly higher quality pseudomasks. The video example (right) is generated by the complete VVitCutLER system and compared with the state-of-the-art unsupervised baseline method CutLER, demonstrating significant improvements in multi-object detection and segmentation.
        The chart (top right) compares bounding box and segmentation APs betwee different annotation methods, while the chart (bottom right) shows the video segmentation performance of our full framework compared to other video models.
    }
    \label{fig:resultall}
\vspace{3mm}
\end{strip}

\begin{abstract}

Unsupervised pixel-level video understanding remains challenging in real-world scenarios, where motion blur, occlusion, and fast object dynamics often cause temporal drift and flickering pseudo-labels.We propose VVitCutLER, an unsupervised framework for video object detection and instance segmentation, which improves the quality of pseudo-labels through temporal consistency. Our core contribution is VitCut, a temporarily stable pseudo-label generator that reduces error accumulation during field degradation through cross-frame region consistency. Meanwhile, VitCut uses a distillation decoder to achieve effective instance mask prediction. Then, based on VitCut, VVitCutLER further integrates cross-frame feature aggregation to enhance video-level robustness. Extensive experiments on standard video benchmarks demonstrate that VVitCutLER significantly improves detection and segmentation performance while reducing temporal instability. These results highlight the importance of temporally consistent supervision for robust pixel-level video understanding. The project code: \url{https://github.com/zhijingLu/VVitCutLER}
\end{abstract}

\section{Introduction}
Video understanding is a fundamental capability for practical applications such as autonomous driving~\cite{hadizadeh2024learnedmultimodalcompressionautonomous}, robotics~\cite{moon2024learningtemporalcuespredicting}, and intelligent surveillance~\cite{fernandeztesta2024distributedintelligentvideosurveillance}. A core requirement in these applications is the reliable localization and segmentation of multiple targets under challenging video conditions of motion blur, occlusion, illumination variations, and scale variations~\cite{zhu2017flowguidedfeatureaggregationvideo, li2019motionguidedattentionvideo}. While supervised detectors and instance segmenters have achieved significant performance in static images~\cite{ren2016fasterrcnnrealtimeobject, he2018maskrcnn, redmon2016lookonceunifiedrealtime, Liu_2016}, and supervised video models further utilize temporal cues (e.g., FGFA ~\cite{zhu2017flowguidedfeatureaggregationvideo}, TubeTK~\cite{pang2020tubetkadoptingtubestrack}), extending these methods to different video domains is hampered by annotation costs. Instance masking of long video sequences is particularly time-consuming, as it requires intensive frame-by-frame annotation.

This has prompted researchers to seek unsupervised video object detection and instance segmentation methods, aiming to learn from unlabeled videos by generating pseudo-labels and using self-training standard architectures. The progress made in unsupervised image object discovery and segmentation in recent years has mainly benefited from self-supervised features and pseudo-label guidance. CutLER~\cite{wang2023cutlearnunsupervisedobject} proposed a self-training pipeline in which foreground candidate boxes are generated by MaskCut based on DINO features and then used to train an instance segmentation model. CuVLER/VoteCut~\cite{arica2024cuvlerenhancedunsupervisedobject} further improved pseudo-masks by integrating multiple self-supervised ViT backbone networks and assigning confidence scores to filter noisy predictions. However, when applied frame by frame to video, these image-based pipelines become fragile. Since the pseudo-labeling process only considers the single-frame situation, the pseudo-box and mask will drift between frames or flicker during fast movement.

A key finding of our work is that the main problem in video detection is not the lack of frame-by-frame detection clues, but the temporal instability of pseudo-labels. Optical flow and time aggregation can be helpful, but when motion estimation is imperfect (for example, fuzzy, non-rigid motion), simply propagating the mask will amplify the error. Therefore, we focus on enhancing the reliability of pseudo-labels through regional-level temporal stability rather than dense mask propagation. In the actual unsupervised setup, the target video does not provide manual annotations, and we utilize the pre-trained model as the fixed prior knowledge. We propose VVitCutLER, an unsupervised and self-trained framework that improves the generation of pseudo-labels and downstream training for unlabeled videos. VVitCutLER consists of two parts: (1) The video pseudo-label module VitCut, which generates temporarily stable target areas and high-quality pseudo-masks; (ii) Self-trained detectors that learn from these pseudo-labels.

The core idea of VitCut is to first generate frame-by-frame candidate target regions using class-agnostic target discovery methods (e.g., VoteCut~\cite{arica2024cuvlerenhancedunsupervisedobject}/ MaskCut~\cite{wang2023cutlearnunsupervisedobject}), and then use motion alignment to stabilize these regions temporally. Specifically, we leverage optical flow to align candidate boxes from adjacent frames and merge them into stable region proposals. This region-level temporal stability reduces flicker and missed detections while avoiding artifacts caused by dense mask deformation.

To generate pseudo-masks efficiently from these stable regions, VitCut employs an offline teacher-student distillation strategy. A powerful segmentation teacher(SAM2) generates masks using full image context, while a lightweight student decoder learns to predict masks from cropped boxes. After distillation, the student replaces the teacher for large-scale pseudo-label generation, significantly reducing inference costs while maintaining quality.

Finally, we use the generated pseudo-labels to train a standard instance segmentation detector (Mask R-CNN~\cite{he2018maskrcnn}) using a self-training mechanism similar to CutLER. To improve robustness to video quality degradation, we introduce Sequence-Level Semantic Aggregation (SELSA)~\cite{wu2019sequencelevelsemanticsaggregation} as a feature aggregation module, leveraging neighboring frames to enhance frame-by-frame representations. This improves consistency in complex frames without requiring identity tracking. 

Experiments on standard video benchmarks show that VVitCutLER outperforms unsupervised baselines in pseudo-label quality and downstream performance, particularly under complex motion and occlusion. These results validate the effectiveness of temporally stable region alignment and efficient pseudo-mask generation for scalable unsupervised video learning. Our main contributions are:\\
\textbf{Region-level temporal stabilization for pseudo-labeling.} We propose a motion-guided region alignment and fusion strategy that stabilizes inter-frame box locations before mask distillation, yielding more temporally consistent pseudo-labels in unlabeled videos.\\
\textbf{Efficient pseudo-mask generation via offline distillation.} We train a lightweight box-conditioned student decoder from scratch using pseudo-supervision from a strong teacher, enabling scalable pseudo-mask generation with low computational overhead.\\
\textbf{Unsupervised self-training with cross-frame feature aggregation.} We train a standard instance segmentation detector on the pseudo-labels and integrate SELSA [46] to aggregate neighboring-frame RoI features, improving robustness to real-world video degradations without identity tracking.

\section{Related Work}
\textbf{Supervised Object Detection and Segmentation} 
Supervised object detection has evolved from two-stage detectors (e.g., the R-CNN series~\cite{girshick2014richfeaturehierarchiesaccurate, ren2016fasterrcnnrealtimeobject} and instance segmentation extensions (e.g., Mask R-CNN\cite{he2018maskrcnn})) to efficient single-stage designs (including SSD~\cite{Liu_2016}, RetinaNet\cite{lin2018focallossdenseobject}, and YOLO\cite{redmon2016lookonceunifiedrealtime}). Then, transformer-based detectors (e.g., DETR~\cite{carion2020endtoendobjectdetectiontransformers} and Deformable DETR~\cite{zhu2021deformabledetrdeformabletransformers}) have further achieved end-to-end prediction through attention mechanisms. While these supervised learning pipelines offer excellent performance, scaling them to long videos remains challenging due to the high cost of dense frame-level annotation; therefore, unlabeled video learning has emerged.\\
\textbf{Unsupervised Object Detection} Unsupervised object detection typically relies on transferable self-supervised representations learned from pre-trained tasks (e.g., MoCo~\cite{he2020momentumcontrastunsupervisedvisual} and DINO~\cite{caron2021emergingpropertiesselfsupervisedvision}), which encode object-centric semantic boundaries. Based on these features, label-free localization methods, including LOST~\cite{siméoni2021localizingobjectsselfsupervisedtransformers} and TokenCut~\cite{guo2022token}, can discover salient objects by analyzing the similarity of image patches/tags, but these methods often bias towards dominant instances. To support multi-object detection, CutLER~\cite{wang2023cutlearnunsupervisedobject} introduces MaskCut (normalized cut) and iterative self-training, while CuVLER(VoteCut)~\cite{arica2024cuvlerenhancedunsupervisedobject} aggregates masks from multiple self-supervised backbone networks and performs confidence-aware refinement. However, these image-centric methods are frame-local: their pseudo-label quality is evaluated frame-by-frame, and they do not explicitly model the temporal uncertainties that are crucial in video.\\
\textbf{Application of Video Object Detection in Image Degradation Cases} Video object detection leverages temporal context to handle motion blur, occlusion, and appearance changes. Early methods concatenate frame-level detection results through temporal post-processing, such as Seq-NMS~\cite{han2016seqnms}, Tubelet Linking~\cite{kang2017tcnn}, and ByteTrack~\cite{zhang2022bytetrackmultiobjecttrackingassociating}. Feature aggregation methods, including FGFA~\cite{zhu2017flowguidedfeatureaggregationvideo} and SELSA~\cite{wu2019sequencelevelsemanticsaggregation}, align and fuse information from adjacent frames to enhance the representation of regions of interest (RoIs). Transformer-based video detectors (such as TransVOD~\cite{he2022transvodendtoendvideoobject} and the YOLOV series ~\cite{shi2023yolovmakingimageobject,shi2024practicalvideoobjectdetection}) integrate spatiotemporal modeling to achieve end-to-end inference. These works emphasize an important principle for handling degraded videos: borrowing contextual information from adjacent frames can compensate for damaged frames. However, most methods are developed under supervised labeling and do not address the reliability issue of pseudo-supervision in unlabeled videos.\\
\textbf{Unsupervised Video Object Detection and Segmentation} Unsupervised video object detection/segmentation utilizes motion cues and temporal consistency, requiring no labels. Early methods used optical flow ~\cite{yang2019unsupervised} or spatiotemporal graphs~\cite{haller2017unsupervisedobjectsegmentationvideo} to identify moving targets, while newer methods employ pseudo-label self-training (e.g., VideoCutLER~\cite{wang2023videocutlersurprisinglysimpleunsupervised}, STC-Seg~\cite{yan2023solve}) or learn video representations through temporal contrast/self-supervised targets (e.g., TCLR~\cite{DAVE2022103406}, VideoMAE~\cite{tong2022videomaemaskedautoencodersdataefficient}).Despite these advancements, real-world video complexity remains a major bottleneck. Recent benchmarks such as MOSE~\cite{ding2023mosenewdatasetvideo} and MOSEv2~\cite{ding2025mosev2challengingdatasetvideo} highlight challenges like severe occlusions and crowded scenes, while MeViS~\cite{ding2023mevislargescalebenchmarkvideo} and MeViSv2~\cite{Ding_2025} emphasize the diversity of motion patterns. Under such conditions, motion and appearance cues become unreliable, and simple propagation/association methods accumulate errors, leading to pseudo-label drift and flickering. This motivates our focus on improving pseudo-label stability through region-level temporal consistency.
\section{Method}
This study presents a novel approach for unsupervised video object detection and segmentation Our method builds upon recent advancements in the field\cite{wang2023cutlearnunsupervisedobject,arica2024cuvlerenhancedunsupervisedobject}. 
\begin{figure*}[ht]
  \centering
  \includegraphics[width=0.7\textwidth]{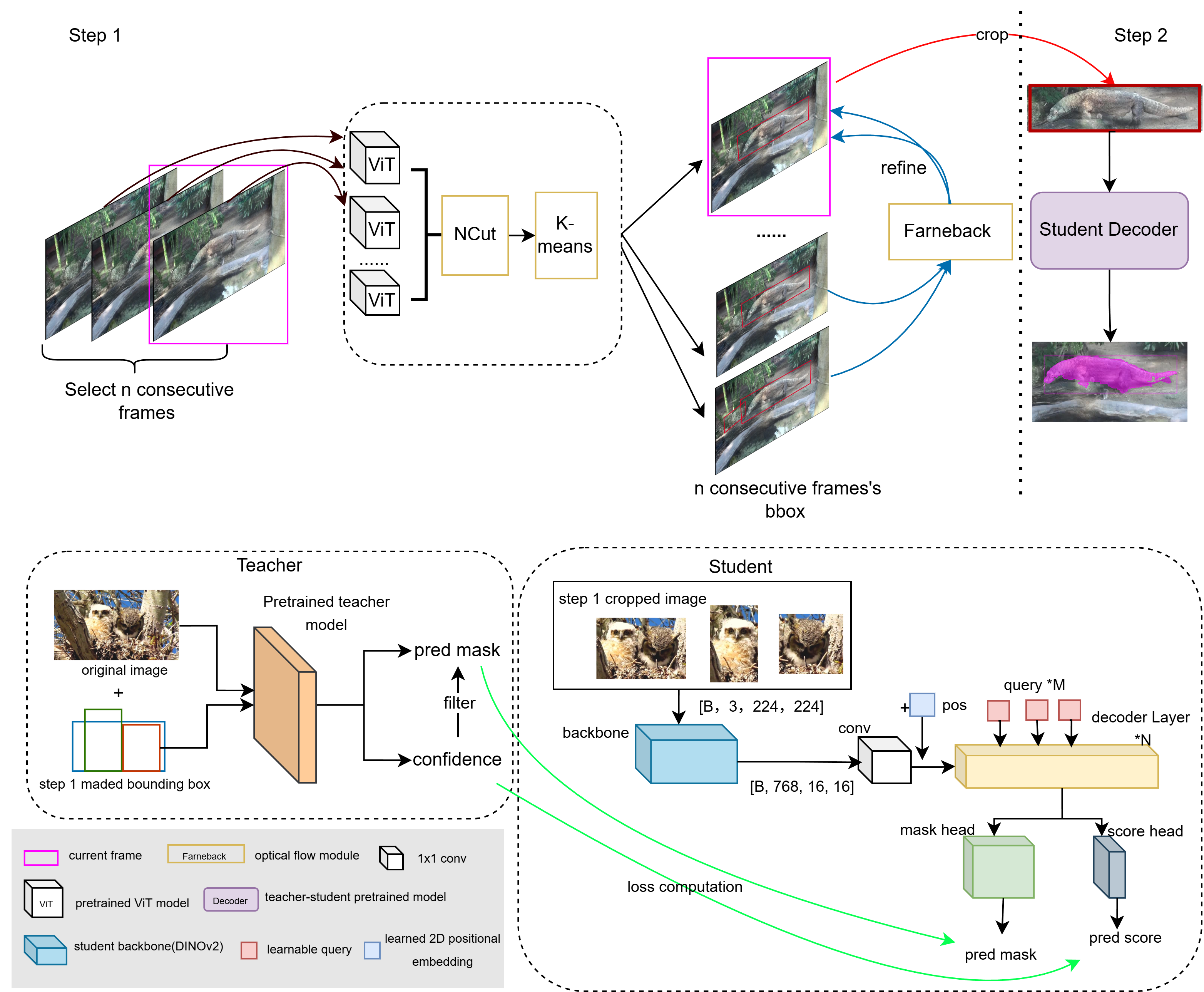}
  \caption{The upper part illustrates the complete two-step process of \textbf{VitCut}.
Step 1: multiple ViT backbones are used to extract features from n consecutive frames, and NCut combined with k-means clustering is applied to obtain multiple object bounding boxes per frame. Using Farneback optical flow, the bounding boxes from the reference frame are aligned and fused with those of the current frame to generate a larger, motion consistent bounding box. Step 2: the cropped region corresponding to the fused bounding box is fed into a pretrained decoder to produce the final object mask. The lower part shows the detailed procedure of the pretrained Teacher–Student framework, which demonstrates how the student decoder is trained. The teacher model provides pseudo masks and confidence scores to filter unreliable samples and supervise the learning of both the mask and score heads in the student decoder.}
  \label{fig:vitcutpipeline}
\end{figure*}

\subsection{VitCut: Pseudo-label Generation for Unlabeled Videos}
VitCut is a pseudo-label generation module used to construct training annotations from unlabeled videos. Given video frames$\{I_t\}$, VitCut outputs frame-level pseudo annotations$\{(\hat{b}_t^{k}, \hat{m}_t^{k})\}$, i.e., bounding boxes and instance masks, which are then used to train the downstream detector/segmenter. \cref{fig:vitcutpipeline} shows an overview of our approach VitCut, which exhibits superior efficiency in processing large datasets and significantly improves model training results. VitCut is mainly divided into two parts, 1) motion-guided temporal alignment and fusion to stabilize candidate regions across neighboring frames, 2) offline teacher–student distillation, where a lightweight box-conditioned student decoder is trained from scratch using teacher-generated pseudo masks, enabling efficient high-quality pseudo-mask generation.

\subsubsection{Candidate Extraction}
\label{sec:candidate}
For each frame $I_t$, we follow VoteCut~\cite{arica2024cuvlerenhancedunsupervisedobject} to generate class-agnostic candidates.
Concretely, VoteCut performs NCut partitioning and k-means clustering on multi-backbone ViT features, and consolidates consistent foreground regions via voting.
We take the voted regions as object candidates and convert them into bounding boxes by computing the tight axis-aligned rectangles, forming the initial set $B_t$.
We keep the top-$K$ candidates per frame after confidence filtering (and optional NMS).

\subsubsection{Temporal Box Stabilization}
\label{sec:temporal}

Frame-level candidate labels extracted independently from videos are often unstable due to motion blur, occlusion, and rapid object movement.

To improve the reliability of pseudo-labels, we stabilize candidate labels at the \emph{box level} by performing motion-guided alignment and fusion within a short time window.\\
\textbf{Motion-guided box alignment.}
For each target frame $I_t$, we use the three preceding and three subsequent frames as references,
$t'\in\{t-3,t-2,t-1,t+1,t+2,t+3\}$ (using available frames near sequence boundaries). This temporal window of ±3 frames is chosen to balance motion context and alignment accuracy: it captures sufficient motion dynamics for typical video frame rates while avoiding excessive drift that can occur with larger windows. For each reference frame $I_{t'}$, we compute dense optical flow from $I_{t'}$ to $I_t$ using Farneb\"ack~\cite{farneback2003two}.
Let $\mathbf{u}_{t'\rightarrow t}(p)=(u(p),v(p))$ denote the flow vector at pixel $p$.
Given a reference box $b_{t'}^k=(x_1,y_1,x_2,y_2)$, we warp it into the coordinate system of frame $t$ by translating the box with the average flow inside the box region:
\begin{equation}
\label{eq:box_warp}
\begin{aligned}
(\bar{u},\bar{v}) &=
\frac{1}{|b_{t'}^k|}
\sum_{p\in b_{t'}^k}\mathbf{u}_{t'\rightarrow t}(p), \\
\tilde{b}_{t'\rightarrow t}^k &=
(x_1+\bar{u},\,y_1+\bar{v},\,x_2+\bar{u},\,y_2+\bar{v}).
\end{aligned}
\end{equation}
where $|b_{t'}^k|$ denotes the number of pixels (or sampled points) inside $b_{t'}^k$.
This produces motion-aligned reference boxes $\tilde{B}_t^{\text{ref}}$ in the coordinate system of the target frame.
The alignment is translation-only and operates at the box level. We adopt this coarse alignment for two reasons: (i) our goal is to stabilize bounding box locations, not pixel-accurate masks, making translation sufficient; (ii) more sophisticated flow models (e.g., RAFT~\cite{teed2020raftrecurrentallpairsfield}) introduce significant computational overhead with marginal benefit for box-level stabilization.

\textbf{Refining current-frame boxes with fused references.}
The fused reference proposals $B_t^{\text{fuse}}$ provide temporally supported regions for frame $t$. We refine the current-frame candidates $B_t$ in a reference-supported manner to suppress spurious detections and recover missed objects.
Specifically, we retain a current-frame box $b\in B_t$ if it overlaps with at least one fused reference box, i.e.,
$\max_{f\in B_t^{\text{fuse}}}\mathrm{IoU}(b,f)\ge 0.6$, forming the filtered set $B_t^{\text{keep}}$.
We then incorporate a fused reference box $f\in B_t^{\text{fuse}}$ if it is not sufficiently covered by any retained current-frame box, i.e.,
$\max_{b\in B_t^{\text{keep}}}\mathrm{IoU}(b,f)< 0.7$, yielding the stabilized proposals $\hat{B}_t$.
We adopt a higher IoU threshold (0.7) for reference grouping/matching to avoid overmerging under flow noise, and a lower threshold (0.6) for retaining current-frame boxes to tolerate alignment inaccuracies and box-tightness variations.\\
\indent Box-level refinement is less sensitive to flow noise than dense mask propagation and avoids error accumulation; the stabilized proposals $\hat{B}_t$ are then used as prompts for pseudo-mask generation (\cref{sec:distill}).

\subsubsection{Student Mask Generator via Offline Distillation}
\label{sec:distill}
To enable efficient pseudo-mask generation, we adopt an offline teacher--student distillation scheme. 
A SAM2 teacher predicts masks and confidence scores from the full image, while a lightweight student is trained from scratch to predict masks from cropped RoIs only. After distillation, the teacher is discarded and the student is used for large-scale pseudo-label generation. 
Given stabilized proposals $\hat{B}_t$ (\cref{sec:temporal}), we crop each RoI (resized to a fixed resolution) and feed it to the student, which outputs a mask logit and a confidence score per proposal. 
The student employs a frozen DINOv2 feature extractor and a lightweight decoder head, and is supervised by the teacher outputs with
\begin{equation}
\label{eq:loss_total}
\mathcal{L}=\mathcal{L}_{\mathrm{seg}}+\mathcal{L}_{\mathrm{score}},
\end{equation}
where $\mathcal{L}_{\mathrm{score}}$ is binary cross-entropy on the confidence, and the segmentation loss is
\begin{equation}
\label{eq:loss_seg}
\mathcal{L}_{\mathrm{seg}}=
0.5\,\mathcal{L}_{\mathrm{BCE}}+
0.3\,\mathcal{L}_{\mathrm{Dice}}+
0.2\,\mathcal{L}_{\mathrm{boundary}}.
\end{equation}

Architectural and training details of the student mask generator (decoder design, head configurations, and optimization settings) are provided in the supplementary material.

\subsection{Video Instance Segmentation}
\begin{figure}[h]
  \centering
   \includegraphics[width=0.9\linewidth]{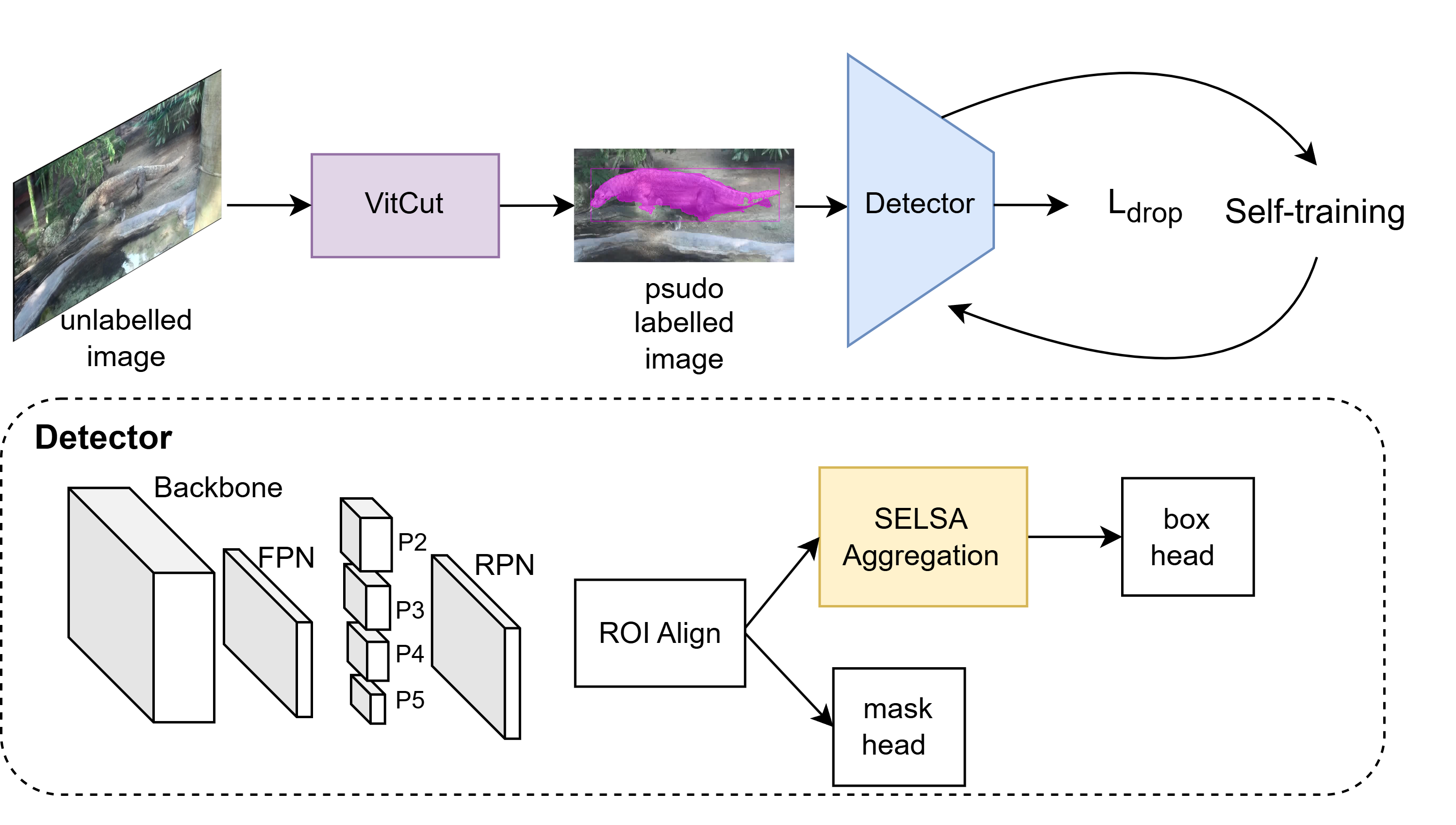} 
   \caption{Overview of \textbf{VVitCutLER}. Unlabeled images are processed by \textbf{VitCut} to produce pseudo masks and boxes, which are used as pseudo labels to train the detector. We adopt self-training, reusing current predictions as pseudo labels for the next round. Within the detector, a SELSA module follows the box head to aggregate temporal features for video learning.}
   \label{fig:detectorpipeline}
\end{figure}
As shown in \cref{fig:detectorpipeline}, we train a video instance segmentation model on unlabeled videos by combining frame-level pseudo supervision with proposal-level temporal feature aggregation.
Our detector is built on Mask R-CNN (ResNet-50 + FPN) and is trained on pseudo annotations produced by VitCut.
Importantly, Mask R-CNN itself remains an image-level instance segmentation model; temporal reasoning is introduced explicitly by a SELSA module that aggregates RoI features across neighboring frames.

\noindent\textbf{SELSA for temporal RoI aggregation.}
During training, each mini-batch corresponds to a short video window sampled from a single video.
For a key frame and its neighboring support frames, we first extract RoI features using the shared backbone and FPN.
We then insert the SELSA~\cite{wu2019sequencelevelsemanticsaggregation} module before the box head to aggregate proposal-level features across frames.
Specifically, SELSA computes semantic affinities between the key-frame RoIs and support-frame RoIs and produces temporally enhanced RoI representations via attention-weighted summation.
The enhanced RoI features are subsequently fed into the classification, box regression, and mask heads for prediction.
This design improves temporal robustness while keeping the prediction heads identical to Mask R-CNN.

\noindent\textbf{Self-training on pseudo labels.}
We follow the CutLER self-training paradigm~\cite{wang2023cutlearnunsupervisedobject}.
In Round-1, the detector is trained on VitCut pseudo labels.
In Round-2, we regenerate pseudo labels using the trained detector predictions and retrain the detector to reduce label noise and improve the representation.
To mitigate noisy pseudo supervision, we employ DropLoss~\cite{wang2023cutlearnunsupervisedobject} during training.
The ResNet-FPN backbone is initialized from DINO pretrained weights and finetuned under pseudo supervision.

\begin{table*}[ht]
\centering
\footnotesize
\renewcommand{\arraystretch}{1.25}
\begin{adjustbox}{width=\linewidth}
\begin{tabular}{
l l 
>{\columncolor{gray!20}}c c 
>{\columncolor{gray!20}}c c 
>{\columncolor{gray!20}}c c 
>{\columncolor{gray!20}}c c 
c}
\toprule
\textbf{Dataset} & \textbf{Method} &
\multicolumn{2}{c}{\textbf{R1 BBox} ↑} &
\multicolumn{2}{c}{\textbf{R1 Segm} ↑} &
\multicolumn{2}{c}{\textbf{R2 BBox} ↑} &
\multicolumn{2}{c}{\textbf{R2 Segm} ↑} &
\textbf{mIoU (\%) ↑}\\
\cmidrule(lr){3-4}\cmidrule(lr){5-6}\cmidrule(lr){7-8}\cmidrule(lr){9-10}
 & & mAP/mAP$_{50}$ & AR(L) & mAP/mAP$_{50}$ & AR(L) & mAP/mAP$_{50}$ & AR(L) & mAP/mAP$_{50}$ & AR(L) &\\
\midrule
\multicolumn{11}{c}{\textbf{DAVIS}}\\
\midrule
VoteCut (U) & & 3.37/6.71 & 21.7 & 2.18/5.04 & 17.3 & 10.29/25.39 & 35.8 & 5.36/11.45 & 23.4 & 3.87\\
VideoCut (U) & & 3.51/7.23 & 27.0 & 2.28/3.98 & 17.3 & 5.21/12.37 & 29.9 & 3.38/6.32 & 19.1 & 2.97\\
\textbf{VitCut (U)} & & 
\textbf{12.49/27.53} & \textbf{36.1} &
\textbf{7.87/14.99} & \textbf{24.6} &
\textbf{14.06/35.03} & \textbf{37.0} &
\textbf{9.79/18.45} & \textbf{23.7} & 
\textbf{4.22}\\
\textit{vs. prev. SoTA} & &
\textcolor{green!50!black}{\textbf{+9.12 }} & \textcolor{green!50!black}{\textbf{+14.4}} &
\textcolor{green!50!black}{\textbf{+5.69 }} & \textcolor{green!50!black}{\textbf{+7.3 }} &
\textcolor{green!50!black}{\textbf{+3.77 }} & \textcolor{green!50!black}{\textbf{+1.2 }} &
\textcolor{green!50!black}{\textbf{+4.43 }} & \textcolor{green!50!black}{\textbf{+0.3 }} &
\textcolor{green!50!black}{\textbf{+0.35 }}\\
\midrule
\multicolumn{11}{c}{\textbf{ImageNetVID}}\\
\midrule
CutSAM (S) & & 11.37/18.58 & 56.8 & -- & -- & 19.91/33.94 & 58.3 & -- & -- & --\\
CutSAM\_2model (S) & & 12.58/18.97 & 57.2 & -- & -- & 15.30/28.71 & 59.4 & -- & -- & --\\
VoteCut (U) & & 12.22/26.39 & 40.0 & -- & -- & 18.34/38.31 & 45.3 & -- & -- & --\\
\textbf{VitCut (U)} & &
\textbf{17.12/37.94} & \textbf{42.9} & -- & -- &
\textbf{19.71/40.30} & 44.3 & -- & -- & --\\
\textit{vs. prev. SoTA} & &
\textcolor{green!50!black}{\textbf{+4.90/+11.55 }} & \textcolor{green!50!black}{\textbf{+2.9 }} &
-- & -- &
\textcolor{green!50!black}{\textbf{+1.37/+1.99 }} & \textcolor{red!50!black}{\textbf{–1 }} &
-- & -- & --\\
\midrule
\multicolumn{11}{c}{\textbf{YouTube-VIS}}\\
\midrule
CutSAM (S) & & 19.56/31.39 & 51.7 & 16.80/28.84 & 46.4 & 23.89/41.63 & 52.1 & 19.16/34.32 & 45.5 & 5.58\\
CutSAM\_2model (S) & & 20.66/32.12 & 53.5 & 17.93/29.34 & 48.1 & 27.74/45.91 & 55.0 & 22.63/39.02 & 47.0 & 5.96\\
VoteCut (U) & & 21.02/39.05 & 45.3 & 14.53/29.79 & 36.5 & 21.21/40.99 & 45.8 & 14.87/30.62 & 37.0 & 6.27\\
VideoCut (U) & & 17.87/32.45 & 45.6 & 12.91/26.23 & 36.5 & 20.48/40.02 & 46.4 & 13.78/29.58 & 37.5 & 5.57\\
\textbf{VitCut (U)} & & 
\textbf{26.85/45.74} & \textbf{50.0} &
\textbf{22.70/40.61} & \textbf{44.5} &
\textbf{28.70/52.23} & \textbf{51.7} &
\textbf{23.55/42.81} & \textbf{44.6} &
\textbf{8.77}\\
\textit{vs. prev. SoTA} & &
\textcolor{green!50!black}{\textbf{+5.83 }} & \textcolor{green!50!black}{\textbf{+4.7 }} &
\textcolor{green!50!black}{\textbf{+8.17 }} & \textcolor{green!50!black}{\textbf{+8.0 }} &
\textcolor{green!50!black}{\textbf{+7.49 }} & \textcolor{green!50!black}{\textbf{+5.9 }} &
\textcolor{green!50!black}{\textbf{+8.68 }} & \textcolor{green!50!black}{\textbf{+7.6 }} &
\textcolor{green!50!black}{\textbf{+2.81 }}\\
\bottomrule
\end{tabular}
\end{adjustbox}
\caption{\textbf{Results on DAVIS, ImageNet-VID, and YouTube-VIS across two self-training rounds.}
R1/R2 denote the first/second round of self-training, respectively.
\textbf{VitCut} is our pseudo-label generator, while \textbf{VideoCut} and \textbf{CutSAM/CutSAM\_2model} are internal baselines implemented by us (details in the supplementary material).
ImageNet-VID provides box annotations only, hence segmentation metrics are unavailable.
(U) indicates unsupervised training and (S) indicates supervised training. VideoCut and CutSAM are internal baselines and are not considered when identifying the best published baseline.}
\label{all_results_combined_cutsam}
\vspace{-4pt}
\end{table*}

\section{Experiment}

This section presents the experimental setup and results for evaluating VVitCutLER.
We report (i) the quality and transferability of VitCut pseudo labels across datasets, and (ii) the downstream video instance segmentation performance obtained by self-training a Mask R-CNN detector augmented with SELSA.
Together, these experiments quantify the contribution of pseudo-label generation and temporal RoI feature aggregation.
\cref{fig:resultall} compares our approach with prior unsupervised baselines.

\subsection{Implementation Details}
\label{sec:impl}
\noindent\textbf{Datasets.}
We evaluate on ImageNet-VID~\cite{russakovsky2015imagenetlargescalevisual}, YouTube-VIS 2021~\cite{nguyen20211stplacesolutionyoutubevos}, and DAVIS 2017~\cite{Pont-Tuset_arXiv_2017}, covering video detection and video instance segmentation under diverse motion and appearance variations.

\noindent\textbf{VitCut stabilization.}
We use a $\pm2$-frame neighborhood for motion-guided box stabilization with Farneb\"ack optical flow (reference $\rightarrow$ target).
IoU thresholds are set to $0.7$ for reference grouping, $0.6$ for keeping current-frame boxes, and $0.7$ for adding unmatched fused references; the minimum group size is 3.

\noindent\textbf{Student decoder distillation.}
We distill a lightweight RoI-based mask generator with DINOv2-Base as a frozen feature extractor and SAM2-Hiera-Large as the teacher.
RoI crops are resized to $224\times224$ and masks are predicted at $64\times64$.
The decoder is trained \emph{from scratch} for 40 epochs using AdamW (lr $2\times10^{-4}$, weight decay $10^{-2}$, batch size 64) with a warmup cosine schedule (5-epoch warmup, restart at epoch 20, min lr $10^{-6}$).
During distillation, we filter teacher supervision by keeping RoIs with confidence $>0.7$.

\noindent\textbf{Detector training.}
We train Mask R-CNN (ResNet-50 + FPN) initialized from DINO weights with a two-round self-training scheme.
Round-1 runs for 160k iterations (batch size 8, lr 0.01) and Round-2 runs for 80k iterations (lr 0.005).
SELSA is inserted before the box and mask heads and operates on RoI features within short video windows sampled from a single video.
DropLoss is used to mitigate noisy pseudo labels.

\subsection{Comparison with State-of-the-Art}
We evaluate VitCut on DAVIS, ImageNet-VID, and YouTube-VIS under a fully unsupervised setting; quantitative results are summarized in \cref{all_results_combined_cutsam}.
We report comparisons against the strongest published baseline in our evaluation (VoteCut), and additionally include our internal variants (VideoCut and CutSAM/CutSAM\_2model) as diagnostic references (details in the supplementary material).
Across all datasets and both self-training rounds, VitCut yields consistent gains over VoteCut in box and mask metrics.
In particular, on DAVIS, VitCut improves R2 segmentation mAP by $+4.43$ over VoteCut; on ImageNet-VID, VitCut reaches 19.71/40.30 mAP/mAP$_{50}$ after Round-2 self-training; and on YouTube-VIS, it achieves the best unsupervised performance with 8.77\% mIoU.
These results indicate that improving pseudo-label quality and incorporating temporal RoI feature aggregation jointly benefit robustness under motion blur and occlusion.
\subsection{Analysis: Pseudo-label Generators}
\begin{table}[t]
\begin{adjustbox}{width=\linewidth}
\centering
\begin{tabular}{l|ll|ll}
\textbf{Method} &
\multicolumn{2}{c|}{\textbf{BBox} $\uparrow$} &
\multicolumn{2}{c}{\textbf{Segm} $\uparrow$}\\
\cmidrule(lr){2-3}\cmidrule(lr){4-5}
 & \textbf{AP} & \textbf{AR} & \textbf{AP} & \textbf{AR} \\
\hline
sam2 &
21.14/33.34 & 58.2 &
17.87/30.39 & 52.2 \\
student decoder &
26.85/45.74 & 50.0 &
22.70/40.61 & 44.5 \\
\textit{vs SAM2} &
\textcolor{green!50!black}{+5.71/+12.40} & \textcolor{red}{-8.2} &
\textcolor{green!50!black}{+4.83/+10.22} & \textcolor{red}{-7.7} \\
\end{tabular}
\end{adjustbox}
\caption{\textbf{YouTube-VIS results.} Comparison of replacing our lightweight decoder with SAM2.}
\label{tab:ytvis21_decoder_replace_sam2}
\vspace{-8pt}
\end{table}
\begin{table}[t]
\begin{adjustbox}{width=\linewidth}
\centering
\begin{tabular}{l|ll|ll}
\textbf{Method} &
\multicolumn{2}{c|}{\textbf{BBox} $\uparrow$} &
\multicolumn{2}{c}{\textbf{Segm} $\uparrow$}\\
\cmidrule(lr){2-3}\cmidrule(lr){4-5}
 & \textbf{AP} & \textbf{AR} & \textbf{AP} & \textbf{AR} \\
\hline
sam2 &
8.90/16.15 & 42.9 &
7.37/10.51 & 29.3 \\
\textbf{student decoder} &
\textbf{12.49/27.53} & \textbf{36.1} &
\textbf{7.87/14.99} & \textbf{24.6} \\
\textit{vs SAM2} &
\textcolor{green!50!black}{+3.59/+11.38} & \textcolor{red}{-6.8} &
\textcolor{green!50!black}{+0.50/+4.48} & \textcolor{red}{-4.7} \\
\end{tabular}
\end{adjustbox}
\caption{\textbf{DAVIS results.} Comparison of replacing our lightweight decoder with SAM2.}
\label{tab:davis_decoder_replace_sam2}
\vspace{-8pt}
\end{table}

\begin{table}[t]
\centering
\small
\begin{adjustbox}{width=0.95\linewidth}
\begin{tabular}{l|cc|ccc}
\toprule
\multirow{2}{*}{\textbf{Method}} &
\multicolumn{2}{c|}{\textbf{YouTubeVIS-2021}} &
\multicolumn{3}{c}{\textbf{DAVIS-2017 (val)}} \\
\cmidrule(lr){2-3}\cmidrule(lr){4-6}
& \textbf{AP} $\uparrow$ & \textbf{AP$_{50}$} $\uparrow$
& \textbf{J\&F} $\uparrow$ & \textbf{J} $\uparrow$ & \textbf{F} $\uparrow$ \\
\midrule

\rowcolor{gray!10} \textit{Early baselines} & & & & & \\
MotionGroup$^*$\cite{yang2021selfsupervised} & 0.2  & 1.1  & --   & --   & --   \\
OCLR$^*$\cite{xie2022segmentingmovingobjectsobjectcentric}       & 1.2  & 4.4  & 39.6 & 38.2 & 41.1 \\
DeepSort baseline$^*$\cite{wojke2017simpleonlinerealtimetracking}      & 10.3 & 23.0 & --   & --   & --   \\
\midrule

\rowcolor{gray!10} \textit{CutLER family (image$\rightarrow$video transfer)} & & & & & \\
CutLER   & 12.8 & 29.2 & --   & --   & --   \\
\textbf{VVitCutLER (ours)}            & \textbf{13.1} & \textbf{31.5} & \textbf{24.35} & \textbf{32.8 }& \textbf{15.9} \\
\midrule
\rowcolor{gray!10} \textit{Other unsupervised VIS frameworks} & & & & & \\
VideoCutLER$^*$\cite{wang2023videocutlersurprisinglysimpleunsupervised}            & 17.1 & 38.9 & 42.4 & 39.2 & 45.6 \\
UVIS$^*$\cite{huang2024uvisunsupervisedvideoinstance}   & 17.5 & 35.6 & --   & --   & --   \\
FlowCut$^*$\cite{sari2025flowcutunsupervisedvideoinstance}      & 18.0 & 37.4 & 43.5 & 41.7 & 45.2 \\
\bottomrule
\end{tabular}
\end{adjustbox}
\caption{\textbf{Unsupervised comparison on YouTubeVIS-2021 and DAVIS-2017.}
For YouTubeVIS-2021, AP/AP$_{50}$ results are reported under each method's original setting (train or val split as specified in the respective paper).
For DAVIS-2017 (val), we report the official J (region), F (boundary), and their mean J\&F metrics.
All comparison results are directly quoted from the corresponding original publications and evaluated under their reported protocols.}
\label{tab:ytvis_davis_joint}
\vspace{-6pt}
\end{table}
\cref{tab:ytvis21_decoder_replace_sam2,tab:davis_decoder_replace_sam2} provide a cross-dataset validation of our choice of pseudo-label generator. On both YouTube-VIS and DAVIS, direct-cue SAM2 consistently favors recall (higher AR), yet yields substantially noisier supervision (lower BBox/Segm scores). In contrast, our student decoder consistently improves pseudo-label quality: on YouTube-VIS, it boosts BBox and Segm scores from 21.14/33.34 and 17.87/30.39 to 26.85/45.74 (+5.71/+12.40) and 22.70/40.61 (+4.83/+10.22), while incurring only moderate AR drops (-8.2/-7.7). The same pattern holds on DAVIS, where VitCut (U) improves BBox and Segm scores to 12.49/27.53 (+3.59/+11.38) and 7.87/14.99 (+0.50/+4.48), despite slightly lower AR (-6.8/-4.7). 

This consistent trade-off indicates that SAM2 becomes overly tolerant to noisy cues from unsupervised candidate boxes: to maintain coverage, it tends to expand masks under imprecise boxes, introducing background leakage and instance confusion in cluttered videos. Our decoder, trained on cropped candidate regions, acts as a task-adaptive annotator that learns tighter instance-aligned masks, producing cleaner supervision signals. Therefore, the evidence across datasets supports that our student decoder is better suited as the default pseudo-label generator: for learning, higher-precision pseudo labels are more beneficial than marginally higher recall.

\begin{table*}[htbp]
\centering
\scriptsize
\renewcommand{\arraystretch}{1.3}
\setlength{\tabcolsep}{4pt}
\begin{adjustbox}{width=\linewidth}
\begin{tabular}{
l
>{\columncolor{gray!20}}c c
>{\columncolor{gray!20}}c c
>{\columncolor{gray!20}}c c
>{\columncolor{gray!20}}c c
c}
\toprule
\textbf{Model} &
\multicolumn{2}{c}{\textbf{Round 1 (BBox) $\uparrow$}} &
\multicolumn{2}{c}{\textbf{Round 1 (Segm) $\uparrow$}} &
\multicolumn{2}{c}{\textbf{Round 2 (BBox) $\uparrow$}} &
\multicolumn{2}{c}{\textbf{Round 2 (Segm) $\uparrow$}} &
\textbf{mFPS} \\
\cmidrule(lr){2-3} \cmidrule(lr){4-5} \cmidrule(lr){6-7} \cmidrule(lr){8-9}
 & mAP / mAP$_{50}$ & AR(L) & mAP / mAP$_{50}$ & AR(L) & mAP / mAP$_{50}$ & AR(L)  & mAP / mAP$_{50}$  & AR(L) & (frames/sec) \\
\midrule
baseline & 30.20 / 50.31 & 64.5 & 39.30 / 52.76 & 65.6 & 46.15 / 78.59 & 68.5 & 55.13 / 77.75 & 69.3 & 12.54 \\
baseline+aggre & \textbf{35.37 / 57.17} & 63.9 & \textbf{42.83 / 58.80} & 64.1 & \textbf{50.99 / 77.71} & 67.1 & 51.38 / 74.24 & 61.7 & 16.96 \\
baseline+aggreall & 33.33 / 51.74 & 62.7 & 28.50 / 48.26 & 61.1 & 45.41 / 72.15 & 64.1 & 38.28 / 63.96 & 62.7 & 10.57 \\
\bottomrule
\end{tabular}
\end{adjustbox}
\caption{Ablation study on different aggregation strategies. Gray columns denote $mAP/mAP_{50}$
 results. We compare the baseline (Mask R-CNN), bbox-only aggregation (\textit{aggre}), and full aggregation (\textit{aggreall}) applied to both bounding boxes and segmentation masks across two training rounds. More detailed analyses and additional ablations are provided in the supplementary ablation section.}
\label{tab:ablation_sam2}
\vspace{-5pt}
\end{table*}

\subsection{VVitcutLER Performance}

\cref{tab:ytvis_davis_joint} summarizes the unsupervised VIS results on the YouTubeVIS-2021 and DAVIS-2017 (val) datasets. In the CutLER-style image-to-video transfer paradigm, VVitCutLER outperforms CutLER on the YouTubeVIS-2021 dataset, achieving AP and AP50 of 13.1 and 31.5, respectively (compared to 12.8/29.2). This improvement is most significant for CNN-based downstream models (e.g., Mask R-CNN style). In these models, our video-aware region optimization and cropping-level supervision yield more consistent hypotheses and better mid-to-low IoU matching, reflected in the larger improvement in AP50.

On the DAVIS-2017 dataset, we observe higher region accuracy, while the improvement in the boundary metric $F$ is smaller, indicating limited contour accuracy. This limitation stems primarily from a lightweight decoder that prioritizes efficiency and stable pseudomask generation over fine-grained boundary modeling. Despite these advances, our approach still lags behind recent Transformer-based integrated unsupervised visual information system (VIS) frameworks (e.g., VideoCutLER and FlowCut), suggesting that more robust global context modeling and query-based aggregation remain complementary directions outside our current transfer flow. In summary, VVitCutLER offers a simple and effective upgrade to CutLER/CNN-style flows, but further bridging the gap may require more expressive (and/or higher resolution) mask decoding and richer global interactions.

\section{Ablation and Efficiency}
\subsection{Aggregation Module}
\label{sec:ablation_aggre}
~\cref{tab:ablation_sam2} ablates our temporal aggregation design: (i) baseline (no aggregation), (ii) baseline+aggre (bbox-only aggregation on RoI features), and (iii) baseline+aggreall (box+mask aggregation).

Baseline+aggre delivers the strongest overall gains. It improves Round-1 BBox $mAP_{50}$ from 50.31 to 57.17 and Round-1 Segm $mAP_{50}$ from 52.76 to 58.80, while also running faster (16.96 vs.\ 12.54 mFPS). This suggests that aggregating RoI features with reliable box cues provides more stable temporal correspondence than using per-frame features alone.

\begin{figure}[t]
\centering
\includegraphics[width=0.6\linewidth]{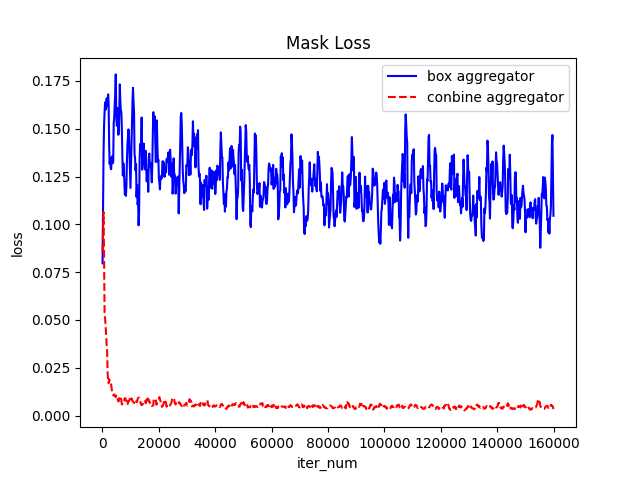}
\caption{Training loss comparison between bbox-only aggregation and full (box+mask) aggregation. Mask aggregation converges faster early but overfits later.}
\label{fig:loss}
\vspace{-6pt}
\end{figure}

In contrast, incorporating mask aggregation is less stable. As shown in~\cref{fig:loss}, adding mask aggregation accelerates early convergence, but the loss later oscillates or rebounds. Consistently, baseline+aggreall yields lower mask accuracy in Table~\cref{tab:ablation_sam2}, likely because noisy pseudo masks propagate inconsistent boundaries across frames. Addressing the trade-off between faster initial learning and late-stage stability is left for future work.

\subsection{Decoder Efficiency Comparison}
\begin{table}[t]
\centering
\small
\begin{adjustbox}{width=0.7\linewidth}
\begin{tabular}{l|cc}
\toprule
\textbf{Model (Backbone)} & \textbf{Params} & \textbf{GFLOPs} \\
\midrule
\rowcolor{gray!10} \textit{ResNet-50 backbone} & & \\
Mask2Former & 44M & 226 \\
MaskDINO & 52M & 286 \\
Decoder for VitCut (ours) & 29.27M & \textbf{97.98} \\
\midrule
\rowcolor{gray!10} \textit{Swin-L backbone} & & \\
Mask2Former & 216M & 868 \\
MaskDINO & 223M & 1326 \\
Decoder for VitCut (ours) & 200.62M & \textbf{371.93} \\
\midrule
\rowcolor{gray!10} \textit{DINOv2-base} & & \\
Decoder for VitCut (ours) & 92.07M & \textbf{512.62} \\
\bottomrule
\end{tabular}
\end{adjustbox}
\caption{\textbf{Model complexity under a unified full-image protocol.}
GFLOPs for our method are converted to the full-image setting (800$\times$1333) using area scaling from 224$\times$224 crops, and reported in a MAC-based convention for consistency with Detectron2-style FLOPs reporting.}
\label{tab:complexity_full}
\end{table}

To evaluate computational efficiency, we compared the number of parameters and GFLOPs of computation across various backbone networks (including ResNet-50, Swin-L, and DINOv2-base) under a unified full-image protocol, and compared them with Mask2Former and MaskDINO. As shown in Table ~\cref{tab:complexity_full}, our VitCut decoder consistently exhibits significantly lower computational complexity.

On the ResNet-50 backbone, our model requires 29.27 million parameters and 9798 GFLOPs of computation, while Mask2Former and MaskDINO require 226 and 286 GFLOPs, respectively. On the Swin-L backbone, our method requires only 37193 GFLOPs of computation, representing a reduction of over 55\% compared to Mask2Former and over 70\% compared to MaskDINO. Even with a DINOv2-based backbone network, VitCut maintains a moderate overall complexity of 512.62 GFLOPs, demonstrating stable scalability even with more powerful feature extractors.

The efficiency improvement stems from replacing the cumbersome pixel-level attention mechanism and multi-scale decoding module with a compact aggregation mechanism based on RoI-aligned features. While this lightweight design may slightly sacrifice segmentation accuracy, it significantly reduces computational cost and model size. In summary, VitCut achieves a good balance between accuracy and efficiency, making it particularly suitable for large-scale or real-time video applications with limited computational resources.

\section{Conclusion}

We propose a framework that integrates pseudo-label generation and temporal feature modeling to extend self-supervised image learning to the video domain. VitCut leverages short-term temporal consistency to stabilize unsupervised pseudo-labels between adjacent frames, providing more reliable region-level supervision for downstream training. Based on these pseudo-labels, we integrate SELSA into a CutLER-style Detector to aggregate cross-frame features and enhance robustness to motion blur and occlusion. Extensive experiments demonstrate significant improvements in both detection and segmentation performance, validating the effectiveness of combining temporally stable annotations with lightweight temporal feature aggregation for unsupervised video understanding.

\noindent\textbf{Limitations:} VitCut improves detection and segmentation \emph{mAP}, but boundary precision is constrained by our lightweight decoder. This design choice favors efficiency and stable pseudo-mask generation, at the cost of reduced capacity for fine contour recovery. Future work will investigate stronger decoders and boundary-aware refinement modules to improve contour fidelity with minimal additional overhead.

\noindent\textbf{Acknowledgments} This research was funded by the EU Horizon Europe program (LUMINOUS, grant number 101135724). The views expressed in this article are solely those of the author and do not represent the views of the European Union.

\clearpage
\setcounter{page}{1}
\maketitlesupplementary

\section{Other Annotation Method}
\label{sec:supplymethods}
In the main paper, we use additional methods referred to as VideoCut and CutSAM for comparison and analysis.
For completeness and reproducibility, we provide their key details below.

\subsection{VideoCut}
\begin{figure*}[ht]
  \centering
  \includegraphics[width=0.8\textwidth]{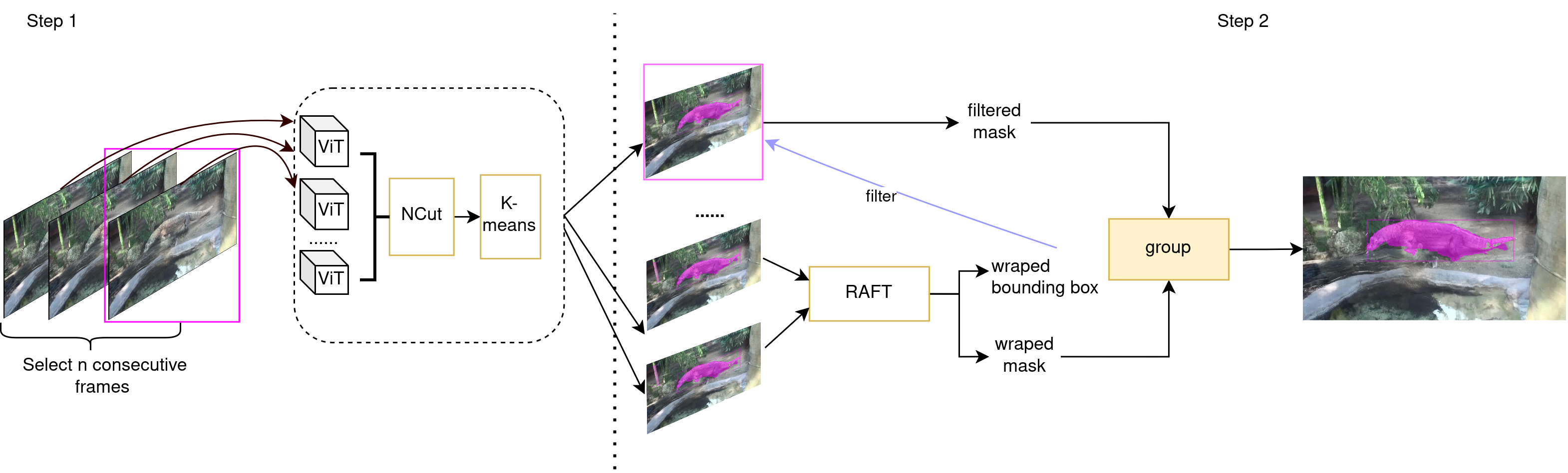}
  \caption{Overview of the proposed VideoCut framework for unsupervised mask extraction. Step 1: Multiple ViT model, NCut, and K-means are applied to each frame to produce initial segmentation masks. Step 2: RAFT is used to estimate dense optical flow between consecutive frames, which warps masks based on temporal motion. The warped masks are compared with the current frame’s masks using IoU, discarding inconsistent or transient objects. Additionally, a background recognition module analyzes average optical flow intensity to detect static regions, which are progressively removed to improve temporal stability.}
  \label{fig:videocutpipeline}
\end{figure*}
VideoCut(see \cref{fig:videocutpipeline}) is an unsupervised variant. Similar to our VitCut method, the first stage of VideoCut follows the VoteCut workflow (see the frame-level unsupervised target discovery section in the main paper). Therefore, we will not elaborate further here.

In the second stage, we first use the RAFT model \cite{teed2020raftrecurrentallpairsfield} to estimate dense optical flow between consecutive frames, which excels at capturing fine-grained motion. Given a current frame $I_t$ and a reference frame $I_{t-1}$, we compute forward optical flow $F_{t-1->t}$ and use it to warp the segmentation mask from the reference frame to the current frame. This generates a predicted mask $M_t^{pred}$ based on temporal consistency.

Next, we compare the warped mask $M_t^{pred}$ with the mask $M_t^{curr}$ directly obtained from the first stage, computing the intersection over union (IoU) of their respective bounding boxes. If the IoU falls below a preset threshold, the object is considered unstable and discarded from the current frame. This filtering step helps eliminate temporally inconsistent transients or false detections.

In addition to motion-guided alignment, we also introduce a background recognition mechanism. For each masked region, we compute the average optical flow intensity. If the intensity is below a certain threshold across multiple consecutive frames, the region is labeled as is\_bg, indicating that it likely belongs to the static background. We then group the masks across frames by their associated bounding box Intersection over Union (IoU) and track the frequency of the is\_bg label. If a region is repeatedly labeled as background, we remove it from all subsequent frames.

\subsection{CutSAM}
\begin{figure*}[ht]
  \centering
  \includegraphics[width=0.8\textwidth]{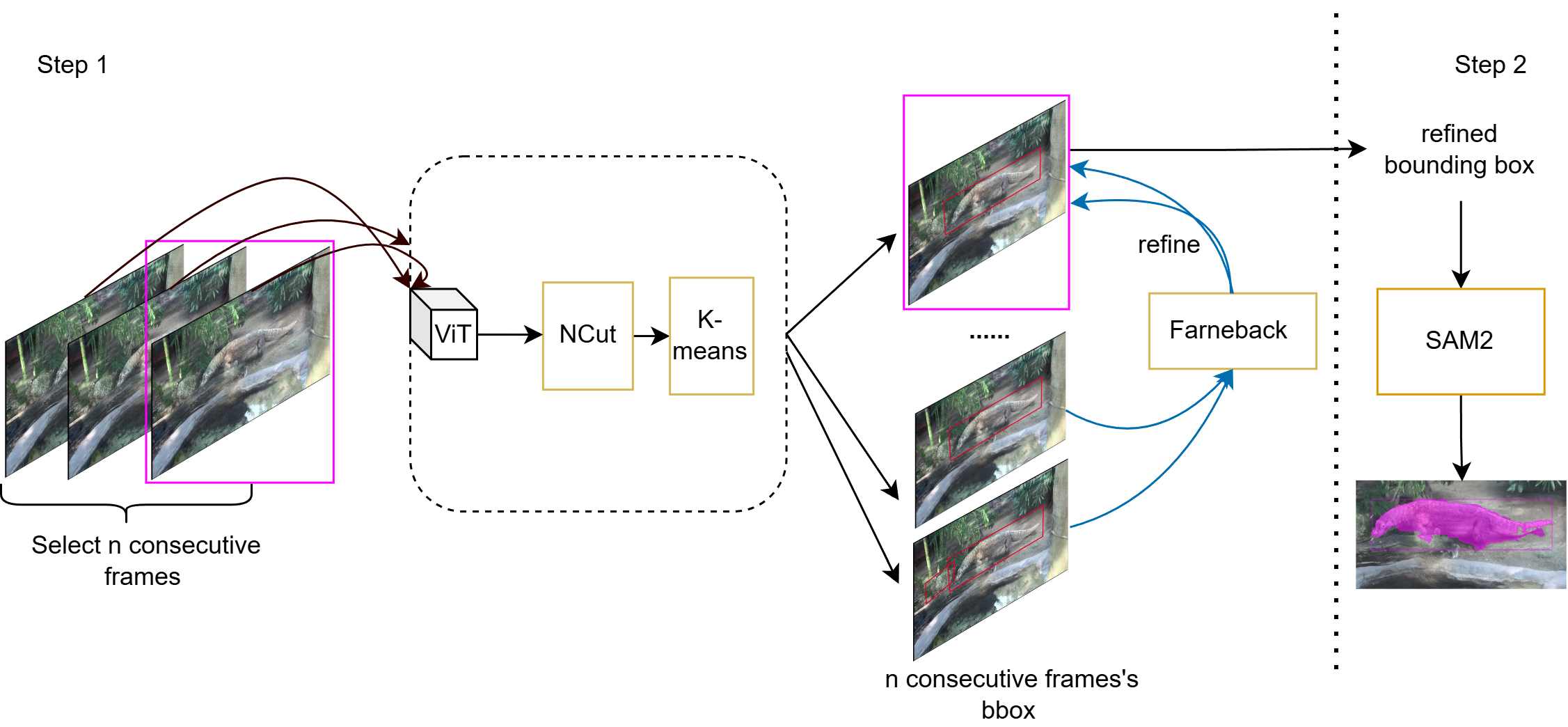}
  \caption{CutSAM's overall architecture. The detected bounding boxes are further clustered to group related regions, and SAM2 is then applied to each cluster to predict a high-quality segmentation mask. This two-step pipeline achieves accurate object segmentation with minimal manual intervention.}
  \label{fig:CutSAM}
\end{figure*}

CutSAM (see \cref{fig:CutSAM}) is a fully supervised variant. Based on the bounding boxes produced in the first stage, we generate object masks using Segment Anything 2 (SAM2) \cite{ravi2024sam2segmentimages}. The overall pipeline follows VideoCut, but with key modifications in the second stage to adapt to the supervised setting.

To enhance the reliability of these bounding boxes, we introduce a temporal consistency improvement process. Specifically, we compute the intersection over union (IoU) of bounding boxes in the reference and target frames. When high IoU overlap is observed, the bounding box from the reference frame is merged into the bounding box of the target frame. This process reduces temporal noise and improves spatial alignment across frames.

Finally, the refined bounding boxes serve as hints to the Segmentation Model 2(SAM2) \cite{ravi2024sam2segmentimages} to generate updated high-quality masks. This step further refines the bounding boxes and segmentation masks, resulting in clearer and more accurate object representations.
\section{Student Mask Generator and Distillation Details}
\label{app:student}

\subsection{Student architecture}
\label{app:student_arch}
The student mask generator predicts an instance mask and a confidence score from a cropped RoI.
Given an RoI crop resized to $224\times224$, we extract features using a frozen DINOv2-base backbone (dinov2-base).
We take the last-layer patch tokens, remove the class token, and reshape them into a $16\times16$ feature map with $C{=}768$ channels.
A lightweight decoder then outputs a mask of resolution $64\times64$ and a scalar confidence score.

\textbf{Decoder configuration.}
We project the DINOv2 features with a $1\times1$ convolution to 256 channels, and apply a Transformer decoder with $L{=}6$ layers, $h{=}4$ attention heads, and feed-forward dimension 512.
The decoder uses $K{=}16$ learnable queries (dimension 256) and a learned 2D positional embedding over the $16\times16$ grid.
For mask prediction, we use a lightweight head consisting of two $3\times3$ Conv-BN-ReLU layers and a final $1\times1$ convolution, followed by bilinear upsampling to $64\times64$.
For confidence prediction, we apply global average pooling on the fused features and use a two-layer MLP to predict a scalar logit.
\subsection{Teacher supervision}
\label{app:teacher}
We use SAM2 as the teacher during offline distillation.
Given a full image and an RoI box, the teacher predicts a binary mask and a confidence score.
Teacher masks are aligned to the student output resolution ($64\times64$) within the RoI region.
After distillation, the teacher is not used during pseudo-label generation; we apply the distilled student checkpoint for efficient inference.
\subsection{Loss}
\label{app:loss}
We follow the loss definition in the main paper (Eq.2and Eq.3).
The student is supervised by the teacher mask and confidence score, where $\mathcal{L}_{\mathrm{score}}$ is binary cross-entropy on the confidence.
The boundary term $\mathcal{L}_{\mathrm{boundary}}$ is implemented by matching Sobel edge responses between the predicted mask probabilities and the teacher mask.
The coefficients $(0.5,0.3,0.2)$ are fixed across all experiments and are chosen to balance the relative contributions of the three terms, with BCE providing stable pixel-level supervision and Dice and boundary losses serving as complementary regularizers for region and contour refinement.

\subsection{Training Details}
\label{app:train}
We train the student decoder \emph{from scratch} and freeze the DINOv2 backbone.
Training is performed with distributed data parallelism.
We use AdamW with learning rate $2\times10^{-4}$ and weight decay $10^{-2}$, batch size 64, for 40 epochs.
We adopt a warmup cosine schedule with a restart: warmup for 5 epochs, cosine decay until epoch 20, and a cosine restart for the remaining epochs, with minimum learning rate $10^{-6}$.
We apply simple data augmentation with random horizontal/vertical flips (each with probability 0.5).

\paragraph{Teacher-score filtering.}
To reduce noisy supervision, we discard RoIs with low teacher confidence during distillation.
Specifically, we keep only samples with teacher score $>0.7$ within each batch.

\section{Qualitative Analysis}

In this section, we present further qualitative results to demonstrate the quality and consistency of VitCut used in this study. We include examples from the datasets described in the main paper, as well as supplementary datasets collected for expanded evaluation. These visualizations highlight the accuracy of the annotation process, the diversity of scenarios, and the robustness of our annotation protocol.

\subsection{Additional Datasets}
\textbf{YouTube-VIS 2019} contains 2,238 videos over 40 categories and serves as a standard benchmark for video instance segmentation under realistic motion and appearance variations.\\
\textbf{OVIS} includes 901 videos across 25 categories and focuses on severe occlusions and crowded scenes, making it a demanding benchmark for occlusion-robust VIS.
\subsection{Qualitative Results of VitCut}

\begin{figure*}[t]
    \centering
    \includegraphics[width=0.9\textwidth]{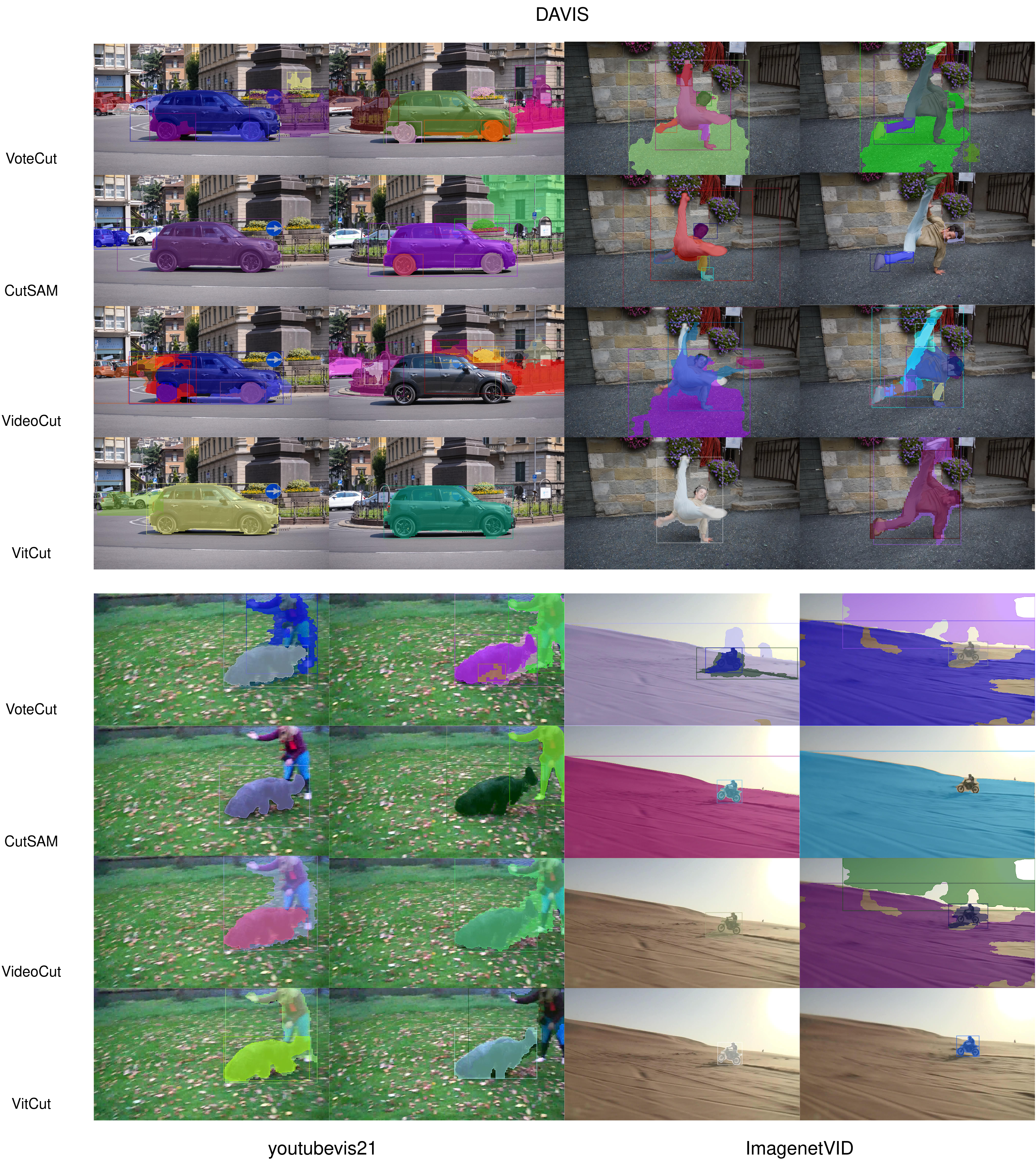}
    \caption{Qualitative visualizations on YouTube-VIS 2021, DAVIS, and ImageNet-VID.}
    \label{fig:qualitative_more_origin}
\end{figure*}
\begin{figure*}[t]
    \centering
    \includegraphics[width=0.9\textwidth]{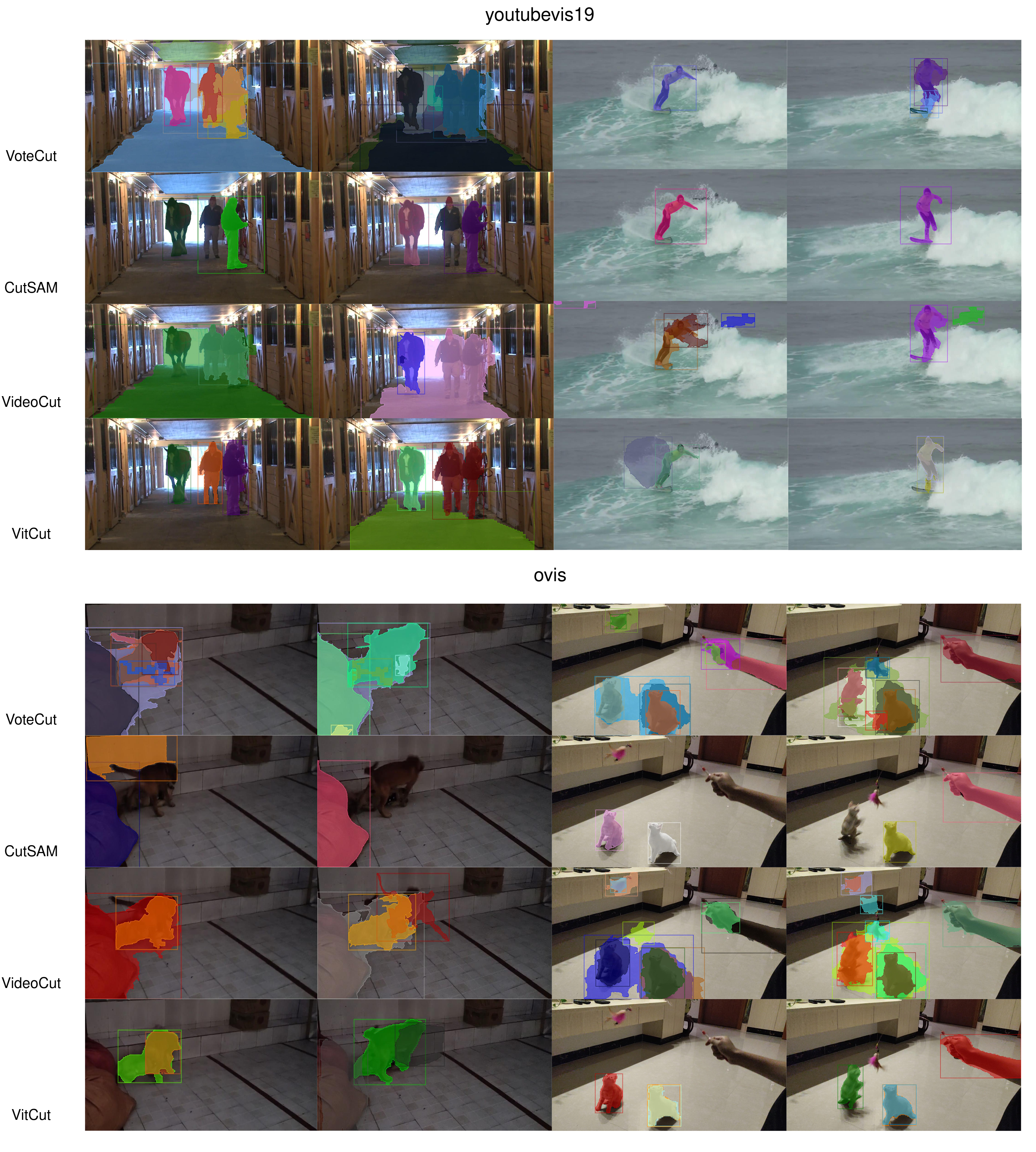}
    \caption{\textbf{Qualitative visualizations on YouTube-VIS 2019 and OVIS.}
    We compare different annotation generation methods on two additional video instance segmentation benchmarks that are not used for training. The results show that \textbf{VitCut} produces more stable and refined masks across diverse scenarios, indicating strong cross-dataset generalization.}
    \label{fig:qualitative_more}
\end{figure*}

The qualitative comparisons in \cref{fig:qualitative_more_origin,fig:qualitative_more} highlight the strengths and limitations of different annotation generation methods. Compared to VideoCut, CutSAM, and VoteCut, our method (\textbf{VitCut}) generates pseudo masks that better match the ground truth and exhibits higher temporal stability across consecutive frames. These results indicate that our Transformer-based design effectively captures long-term dependencies and accommodates changes in object appearance over time.

Nevertheless, a noticeable gap remains between VitCut and the supervised CutSAM baseline. In high-resolution examples, the boundaries produced by VitCut can be less smooth and may miss fine contour details. This is likely due to the combination of (i) the lightweight decoder design, which prioritizes efficiency, and (ii) the limited granularity and noise in the pseudo-annotation training signal, making it challenging to recover highly detailed object contours.

Among the baselines, VideoCut shows the most significant degradation. Its mask quality is strongly affected by optical flow errors, especially under strong motion blur, occlusion, rapid displacement, or non-rigid deformation. In these scenarios, distorted masks can propagate incorrect information across frames, leading to shape drift or severe distortion. This suggests that optical flow-based cues are better used as supplementary signals rather than a fully reliable basis for mask propagation.

Despite these limitations, the results on additional datasets in \cref{fig:qualitative_more} demonstrate that VitCut generalizes well to benchmarks such as YouTube-VIS 2019 and OVIS. Even without training on these datasets, the method consistently generates stable and coherent instance masks, highlighting the robustness of the proposed annotation pipeline.

\subsection{Qualitative Results of VVitCutLER}
\begin{figure*}[t]
    \centering
    \includegraphics[width=0.9\textwidth]{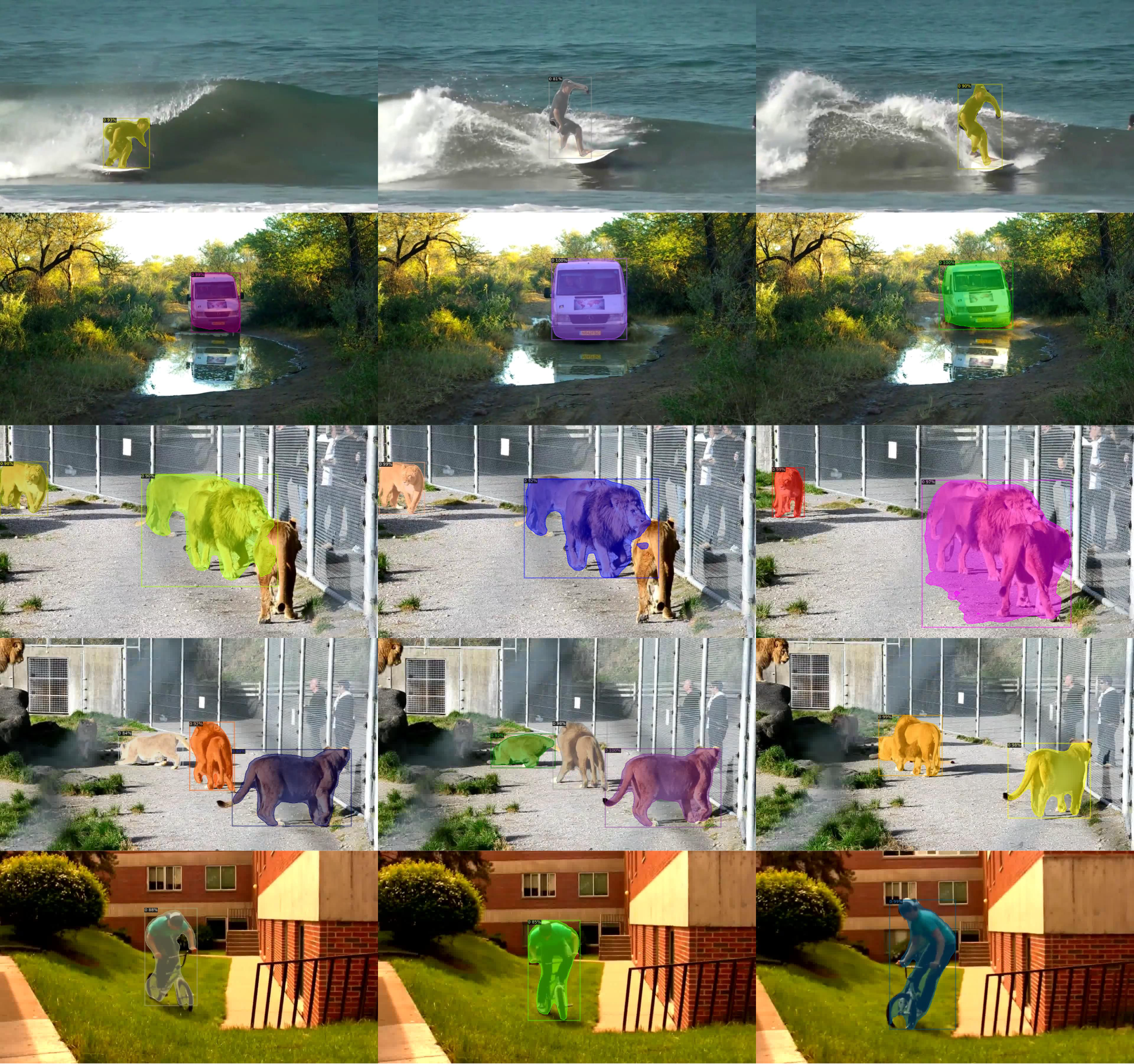}
    \caption{Qualitative visualizations of VVitCutLER on YouTube-VIS 2021 and ImageNet-VID.}
    \label{fig:qualitative_vvitcutler}
\end{figure*}

Beyond generating high-quality pseudo masks for individual instances, the complete \textbf{VVitCutLER} pipeline can handle complex multi-object scenes. As shown in \cref{fig:qualitative_vvitcutler}, the framework can identify multiple objects and maintain mask consistency over time, even under notable changes in appearance and motion.

However, since the pipeline is entirely unsupervised, certain challenging cases remain. When objects move very close to each other or share similar visual patterns, the model may confuse identities or merge them into a single instance. This limitation mainly stems from the absence of explicit supervision signals, which makes it difficult to robustly separate instances in highly cluttered or interactive scenes.

\section{decoder performance}
To evaluate downstream performance under pseudo-label supervision, we report results on ImageNet-VID and YouTube-VIS across two rounds of self-training (R1/R2).
All experiments are conducted in a CNN detector architecture: we adopt Mask R-CNN with a ResNet-50+FPN backbone, and incorporate temporal information only at the RoI-feature level via SELSA while keeping the prediction heads unchanged.
We compare VVitCutLER against published unsupervised baselines, including CutLER~\cite{wang2023cutlearnunsupervisedobject} and CutvLER~\cite{arica2024cuvlerenhancedunsupervisedobject}, under the same evaluation protocol.
\begin{table}[t]
\begin{adjustbox}{width=\linewidth}
\centering
\begin{tabular}{l|l|ll|ll|ll|l}
\textbf{Annotation Type} & \textbf{Model} &
\multicolumn{2}{c|}{R1 mAP$_{50}$ ↑} &
\multicolumn{2}{c|}{R2 mAP$_{50}$ ↑} &
\multicolumn{2}{c|}{AR(L) ↑} &
mFPS ↑\\
\cmidrule(lr){3-4}\cmidrule(lr){5-6}\cmidrule(lr){7-8}
 &  & BBox & Segm & BBox & Segm & R1 & R2 &  \\
\hline
\multirow{4}{*}{Unsupervised}
 & CutLER\cite{wang2023cutlearnunsupervisedobject}      & 28.43 & - & 32.46 & - & 21.1 & 26.5 & 16.68 \\
 & CutvLER\cite{arica2024cuvlerenhancedunsupervisedobject}     & 28.95 & - & 33.96 & - & 23.2 & 29.6 & 16.1 \\
 & VVitCutLER  & \textbf{38.26} & - & \textbf{43.68} & - & 37.5 & 43.7 & 14.93 \\
 & \textit{vs CutLER}
    & \textcolor{green!50!black}{+9.83} & -
    & \textcolor{green!50!black}{+11.22} & -
    & \textcolor{green!50!black}{+16.4} & \textcolor{green!50!black}{+17.2}
    & \textcolor{red}{-1.75} \\
\end{tabular}
\end{adjustbox}
\caption{\textbf{ImageNetVID results under the pseudo-label (unsupervised) setting across Round 1 (R1) and Round 2 (R2).}
We report mAP$_{50}$ and AR(L) for each round, and mFPS for efficiency. Since ImageNetVID is detection-only, the segmentation columns are not applicable (shown as ``-’’).
Green numbers denote absolute gains over the CutLER baseline.}
\label{tab:imagenetvid_results}
\vspace{-8pt}
\end{table}

\begin{table}[t]
\begin{adjustbox}{width=\linewidth}
\centering
\begin{tabular}{l|l|llllllll}
\textbf{Annotation Type} & \textbf{Model} &
\multicolumn{2}{c}{R1 mAP$_{50}$ ↑} &
\multicolumn{2}{c}{R2 mAP$_{50}$ ↑} &
\multicolumn{2}{c}{AR(L) ↑} &
mFPS ↑\\
\cmidrule(lr){3-4}\cmidrule(lr){5-6}\cmidrule(lr){7-8}
 &  & BBox & Segm & BBox & Segm & R1 & R2 &  \\
\hline
\multirow{3}{*}{Unsupervised} 
 & CutLER & 35.21 & 28.54 & 39.04 & 29.03 & 28.2 & 33.8 & 18.98 \\
 & CutvLER & 36.71 & 29.01 & 39.94 & 30.93 & 29.2 & 34.2 & 19.1 \\
 & VVitCutLER & \textbf{37.02} & \textbf{31.43} & \textbf{41.12} & \textbf{32.48} & \textbf{35.2} & \textbf{39.3} & 14.32 \\
 & \textit{vs CutLER} & \textcolor{green!50!black}{+1.81} & \textcolor{green!50!black}{+2.89} & 
 \textcolor{green!50!black}{+2.08} & \textcolor{green!50!black}{+3.45} &
 \textcolor{green!50!black}{+7.0} & \textcolor{green!50!black}{+5.5} &
 \textcolor{red}{-4.66} \\
\end{tabular}
\end{adjustbox}
\caption{\textbf{YouTube-VIS results under the pseudo-label (unsupervised) setting across Round 1 (R1) and Round 2 (R2).}
We report both bounding-box and segmentation mAP$_{50}$, as well as AR(L) for each round, and mFPS for efficiency.
Green numbers denote absolute gains over the CutLER baseline.}
\label{tab:ytvis_results}
\vspace{-8pt}
\end{table}
As shown in \cref{tab:imagenetvid_results,tab:ytvis_results}, VVitCutLER consistently improves over prior baselines on both datasets and in both self-training rounds.
On ImageNet-VID (detection-only), VVitCutLER increases mAP$_{50}$ from 28.43/32.46 (CutLER) to 38.26/43.68 in R1/R2, and yields substantial improvements in AR(L) (+16.4 and +17.2 for R1 and R2, respectively).
On YouTube-VIS, VVitCutLER improves both bounding-box and segmentation accuracy, reaching 37.02/31.43 (BBox/Segm) in R1 and 41.12/32.48 in R2, while also achieving higher AR(L) in both rounds.
The reduction in mFPS reflects the expected accuracy--speed trade-off introduced by temporal RoI feature aggregation.
Overall, these results demonstrate that VVitCutLER provides stronger downstream detection and segmentation performance under unsupervised self-training.

\section{Ablation}
\subsection{Feature Selection}
To analyze the impact of candidate box filtering on our detector, we evaluated different Top-K settings for selecting RPN candidate boxes. Instead of using a fixed confidence threshold, we retained the top K bounding box predictions sorted by RPN confidence score. We experimented with $K \in \{30, 100, 120, 150, 200\}$ and calculated the average recall (AR) on the training set.

\begin{figure}[htbp]
  \centering
  \includegraphics[width=\linewidth]{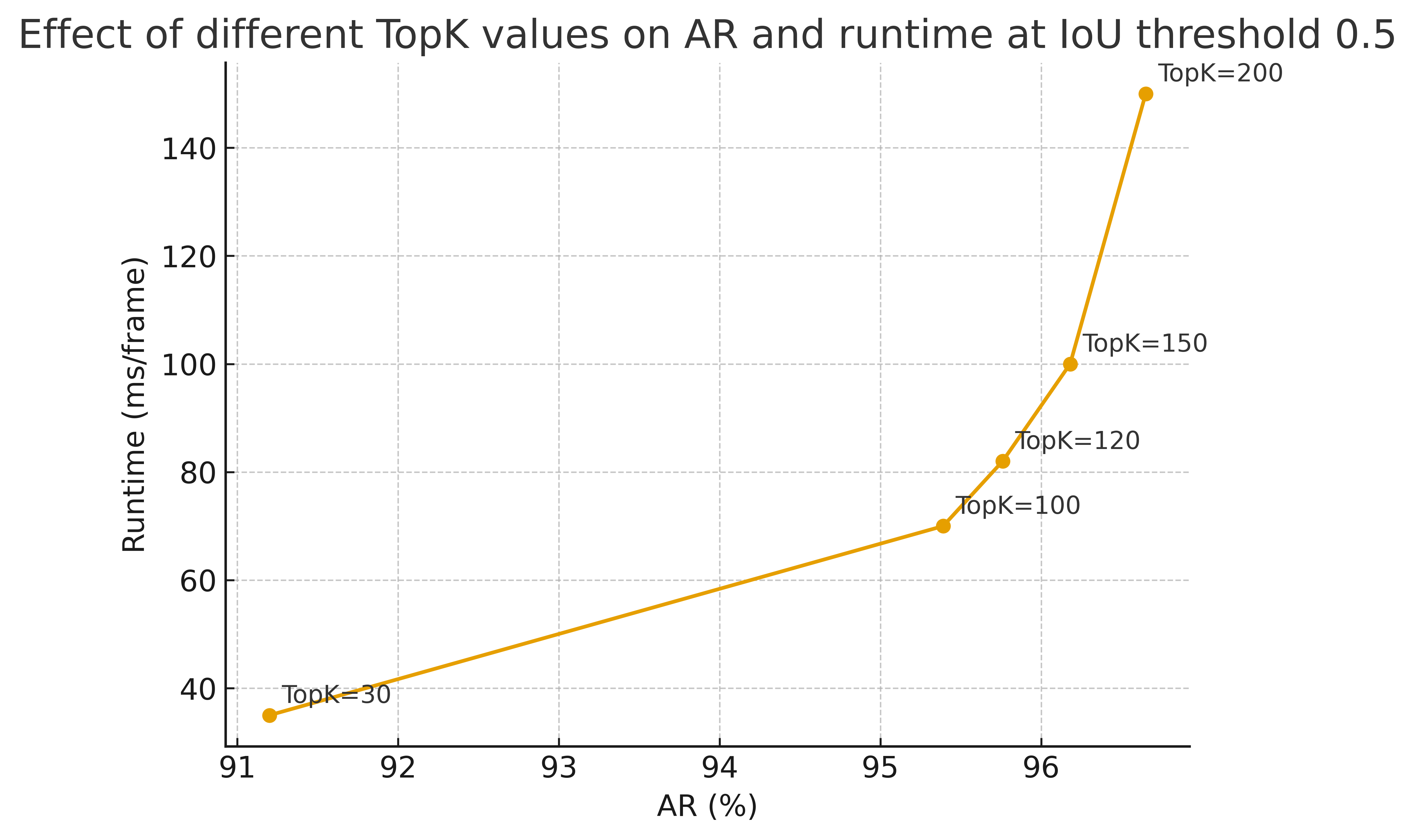}
  \caption{\textbf{Effect of different TopK values on AR and runtime at IoU threshold 0.5.}
TopK=150 achieves the best balance between accuracy and efficiency.}
  \label{fig:topk}
\end{figure}

As shown in \cref{fig:topk}, while increasing the K value consistently improves the AR value, indicating that retaining more candidate boxes allows the detector to capture more potential foreground regions, the runtime also increases almost linearly. This is because a larger K value requires subsequent stages (including feature extraction, classification, and mask prediction) to process more candidate boxes, directly increasing the computational cost per frame.

From an efficiency perspective, Top-K = 150 achieves the best balance between accuracy and cost, achieving an AR value of 96.18\% within a 100ms runtime per frame. While increasing the K value to 200 provides a 0.47\% improvement in AR value, the runtime increases by 50\%, resulting in diminishing returns. For video applications, maintaining low frame latency is crucial, making this additional overhead unacceptable.

In conclusion, selecting the top 150 detector candidates achieves an effective balance between recall and efficiency and serves as the default configuration for our system.

\subsection{Aggregation on Cascade R-CNN}
To evaluate whether our aggregation strategy can be generalized to more complex detector architectures, we further experimented on Cascade R-CNN\cite{cai2017cascadercnndelvinghigh}, a multi-stage detection framework with iterative feature refinement. We integrated the aggregator module at different stages and conducted evaluations.

\begin{figure}[htbp]
  \centering
  \includegraphics[width=\linewidth]{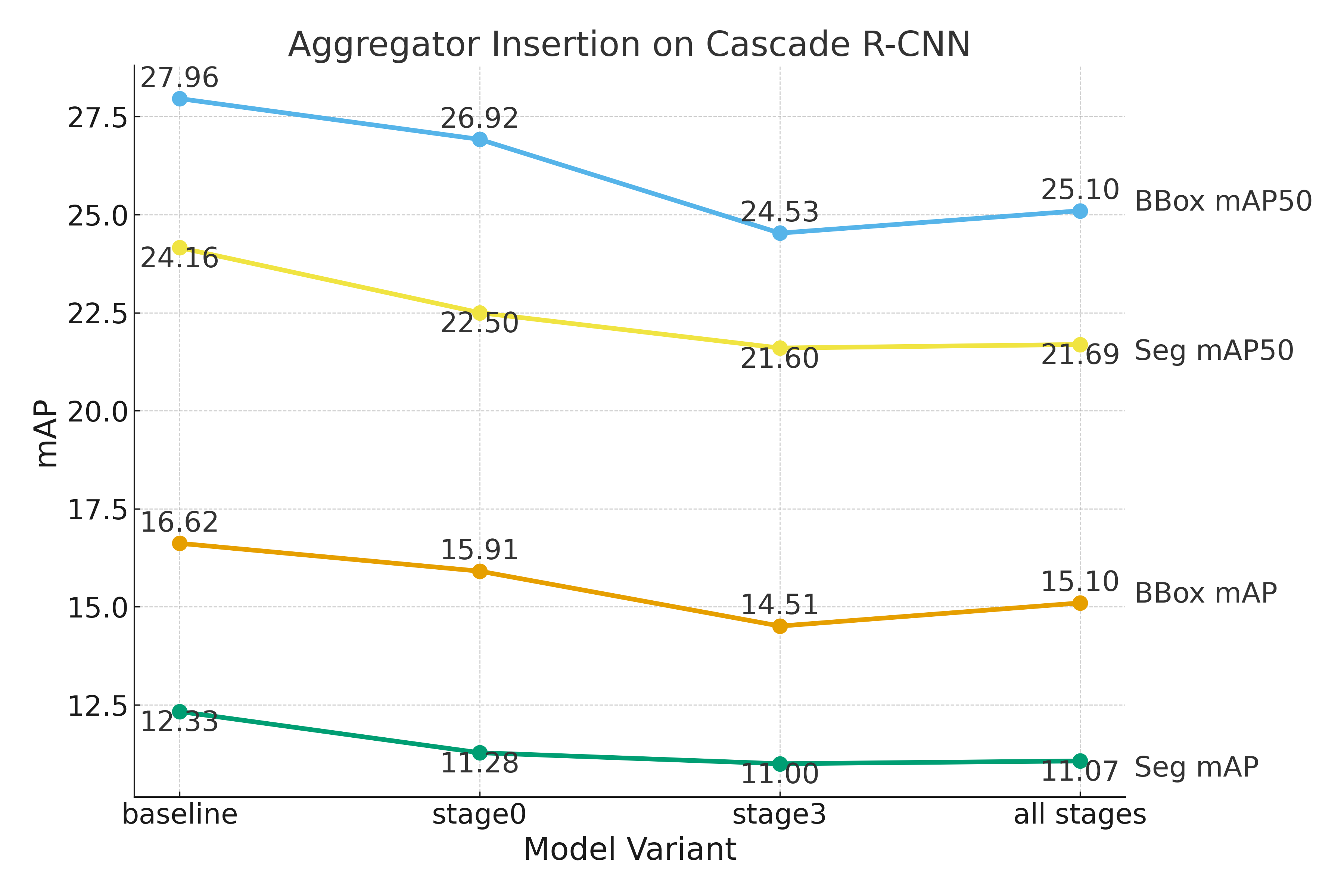}
  \caption{\textbf{Effect of inserting the aggregator module into different stages of Cascade R-CNN on YouTube-VIS.} 
The baseline achieves the best performance, while inserting the aggregator at various stages results in performance degradation.}
  \label{fig:cascadeablation}
\end{figure}

As shown in \cref{fig:cascadeablation}, none of the fusion strategies improved upon the unmodified Cascade R-CNN baseline model. The baseline model still achieved the best overall performance (16.62\% bounding box mAP and 12.33\% segmentation mAP), while the insertion aggregator typically reduced accuracy—especially in stage 3 and all-stage settings.

This phenomenon can be attributed to the multi-stage refinement mechanism of Cascade R-CNN. Each stage relies on clean and progressively refined features; therefore, injecting temporally aggregated features that may contain noise or redundancy can disrupt the refinement process, leading to suboptimal predictions. In contrast, simpler single-stage or two-stage detectors are more tolerant of temporal fusion and can benefit from our aggregation design.

\subsection{Effect of Teacher Model Choice}

To explore how different teacher models affect the quality of pseudo-labels, we compared three candidate models: DINOv2-base, DINOv2-large, and SAM2.

We evaluated each teacher model using mean intersection-over-union (mIoU) and rating accuracy, both of which reflect the quality and reliability of pseudo-labels.
\begin{table}[ht]
\begin{adjustbox}{width=\linewidth}
\centering
\begin{tabular}{l|l|llll}
\textbf{Teacher Category} & \textbf{Teacher Model} & \textbf{mIoU} ↑ & \textbf{Score Acc.} ↑ & & \\
\hline
\multirow{3}{*}{Teacher Models} 
 & DINOv2-base & 0.6644 & 0.9973 & & \\
 & DINOv2-large & 0.3183 & 0.9988 & & \\
 & SAM2 & \textbf{0.6889} & \textbf{1.0000} & & \\
\end{tabular}
\end{adjustbox}
\caption{\textbf{Comparison of pseudo-label quality produced by different teacher models.}
SAM2 achieves the highest segmentation quality (mIoU) and perfect score accuracy, demonstrating its advantages as a teacher model for pseudo-label generation.}
\label{tab:teacher-model-performance}
\vspace{-8pt}
\end{table}
As shown in \cref{tab:teacher-model-performance}, SAM2 achieved the highest mIoU value (0.6889), approximately 3.7\% higher than DINOv2-base and significantly better than DINOv2-large (0.3183). This indicates that SAM2 generates more accurate and finer-grained segmentation masks, providing higher-quality supervision information.

While all teacher models exhibited high score accuracy, SAM2 achieved a perfect score of 1.0000, demonstrating stronger stability and consistency. DINOv2-large achieved slightly higher score accuracy than DINOv2-base, but its segmentation quality dropped sharply, making it unsuitable for pseudo-label generation.

Overall, SAM2 strikes the best balance between segmentation accuracy and label reliability. Its segmentation-oriented design enables it to generate fine-grained, high-quality masks, making it the most effective teacher model in our student training pipeline.

{
    \small
    \bibliographystyle{ieeenat_fullname}
    \bibliography{main}
}


\end{document}